\documentclass[runningheads]{llncs}
\usepackage{graphicx}
\usepackage{amsmath,amssymb} 
\usepackage{color}

\usepackage{algorithm}
\usepackage{algorithmicx}
\usepackage{subfigure}
\usepackage{algpseudocode}
\usepackage{array}
\begin{document}
\title{Fine-Grained Visual Categorization using Meta-Learning Optimization with Sample Selection of Auxiliary Data} 

\titlerunning{Fine-Grained Visual Categorization using Meta-Learning Optimization}
%
\author{Yabin Zhang \and
Hui Tang \and
Kui Jia\thanks{Corresponding author}}

%
\authorrunning{Yabin Zhang, Hui Tang and Kui Jia}
%

\institute{School of Electronic and Information Engineering,\\
	South China University of Technology, Guangzhou, China\\
	\email{ \{zhang.yabin,eehuitang\}@mail.scut.edu.cn, kuijia@scut.edu.cn}
}

\maketitle              
\begin{abstract}
Fine-grained visual categorization (FGVC) is challenging due in part to the fact that it is often difficult to acquire an enough number of training samples. To employ large models for FGVC without suffering from overfitting, existing methods usually adopt a strategy of pre-training the models using a rich set of auxiliary data, followed by fine-tuning on the target FGVC task. However, the objective of pre-training does not take the target task into account, and consequently such obtained models are suboptimal for fine-tuning. To address this issue, we propose in this paper a new deep FGVC model termed MetaFGNet. Training of MetaFGNet is based on a novel regularized meta-learning objective, which aims to guide the learning of network parameters so that they are optimal for adapting to the target FGVC task. Based on MetaFGNet, we also propose a simple yet effective scheme for selecting more useful samples from the auxiliary data. Experiments on benchmark FGVC datasets show the efficacy of our proposed method.

\end{abstract}
\section{Introduction}
\label{SecIntro}

Fine-grained visual categorization (FGVC) aims to classify images of subordinate object categories that belong to a same entry-level category, e.g., different species of birds \cite{cub2002010,cub2002011,birdsnap} or dogs \cite{stanforddog}. The visual distinction between different subordinate categories is often subtle and regional, and such nuance is further obscured by variations caused by arbitrary poses, viewpoint change, and/or occlusion. Subordinate categories are leaf nodes of a taxonomic tree, whose samples are often difficult to collect. Annotating such samples also requires professional expertise, resulting in very few training samples per category in existing FGVC datasets \cite{cub2002011,stanforddog}. FGVC thus bears problem characteristics of few-shot learning.

Most of existing FGVC methods spend efforts on mining global and/or regional discriminative information from training data themselves. For example, state-of-the-art methods learn to identify discriminative parts from images of fine-grained categories either in a supervised \cite{partrcnn,spda} or in a weakly supervised manner \cite{twolevelattention,racnn,dvan,partbased,opaddl,macnn}. However, such methods are approaching a fundamental limit since only very few training samples are available for each category. In order to break the limit, possible solutions include identifying auxiliary data that are more useful for (e.g., more related to) the FGVC task of interest, and also better leveraging these auxiliary data. These solutions fall in the realm of domain adaptation or transfer learning \cite{TransferLearningSurvey}.

A standard way of applying transfer learning to FGVC is to fine-tune on the target dataset a model that has been pre-trained on a rich set
of auxiliary data (e.g., the ImageNet \cite{imagenet}). Such a pre-trained model learns to encode (generic) semantic knowledge from the auxiliary data, and the combined strategy of pre-training followed by fine-tuning alleviates the issue of overfitting. However, the objective of pre-training does not take the target FGVC task of interest into account, and consequently such obtained models are suboptimal for transfer.

\begin{figure*}[t]
\begin{center}
\includegraphics[width=0.95\linewidth]{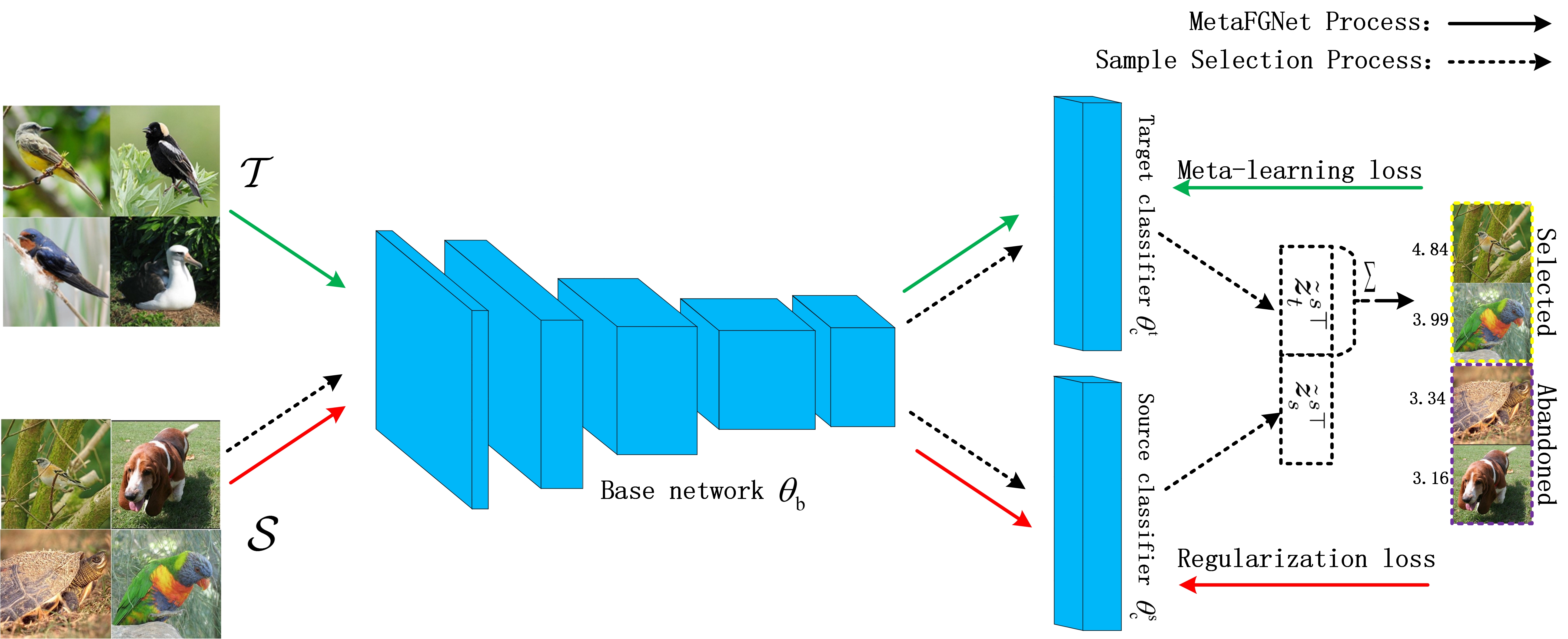}
\end{center}
\caption{Illustrations of MetaFGNet with regularized meta-learning objective (solid line) and the process of sample selection from auxiliary data (dashed line).}
\label{fig:MetaFGNet}
\end{figure*}

Inspired by recent meta-learning methods \cite{maml,matching,metalstm} for few-shot learning, we propose in this paper a new deep learning method for fine-grained classification. Our proposed method is based on a novel \textit{regularized meta-learning objective} for training a deep network: the regularizer aims to learn network parameters such that they can encode generic or semantically related knowledge from auxiliary data; the meta-learning objective is designed to guide the process of learning, so that the learned network parameters are optimal for adapting to the target FGVC task. We term our proposed FGVC method as MetaFGNet for its use of the meta-learning objective. Figure \ref{fig:MetaFGNet} gives an illustration. Our method can effectively alleviate the issue of overfitting, as explained in Section \ref{SecMetaFGNet}.

An important issue to achieve good transfer learning is that data in source and target tasks should share similar feature distributions \cite{TransferLearningSurvey}. If this is not the case, transfer learning methods usually learn feature mappings to alleviate this issue. Alternatively, one may directly identify source data/tasks that are more related to the target one. In this work, we take the later approach and propose a simple yet very effective scheme to select more useful samples from the auxiliary data. Our scheme is naturally admitted by MetaFGNet, and only requires a forward computation through a trained MetaFGNet for each auxiliary sample, which contrasts with a recent computationally expensive scheme used in \cite{jointfinetune}. In this work, we investigate ImageNet \cite{imagenet}, a subset of ImageNet and a subset of L-Bird \cite{l-bird} as the sets of auxiliary data. For the L-Bird subset, for example, our scheme can successfully remove noisy, semantically irrelevant images. Experiments
on the benchmark FGVC datasets of CUB-200-2011 \cite{cub2002011} and Stanford Dogs \cite{stanforddog} show the efficacy of our proposed MetaFGNet with sample selection of auxiliary data. Our contributions are summarized as follows.
\begin{itemize}
\item We propose a new deep learning model, termed MetaFGNet, for fine-grained classification. Training of MetaFGNet is based on a novel \textit{regularized meta-learning objective}, which aims to guide the learning of network parameters so that they are optimal for adapting to the target FGVC task (cf. Section \ref{SecMetaFGNet}).

\item Our proposed MetaFGNet admits a natural scheme to perform sample selection from auxiliary data. Given a trained MetaFGNet, the proposed scheme only requires a forward computation through the network to produce a score for each auxiliary sample (cf. Section \ref{SecSampleSelection}). Such scores can be used to effectively select semantically related auxiliary samples (or remove noisy, semantically irrelevant ones).

\item We present intensive comparative studies on different ways of leveraging auxiliary data for the target FGVC task. Experiments on the benchmark CUB-200-2011 \cite{cub2002011} and Stanford Dogs \cite{stanforddog} datasets also show the efficacy of our proposed method. In particular, our result on Stanford Dogs is better than all existing ones with a large margin. Based on a better auxiliary dataset, our result on CUB-200-2011 is better than those of all existing methods even when they use ground-truth part annotations (cf. Section \ref{SecExp}).
\end{itemize}

\section{Related works}
In this section, we briefly review recent fine-grained classification methods, in particular those aiming for better leveraging auxiliary data, and also meta learning methods for the related problem of few-shot learning. We present brief summaries of these methods and discuss their relations and differences with our proposed one.

\noindent\textbf{Fine-grained visual categorization} State-of-the-art FGVC methods usually follow the pipeline that first discovers discriminative local parts from images of fine-grained categories, and then utilizes the discovered parts for classification. For example, Lam \textit{et al.} \cite{hsnet} search for discriminative parts by iteratively evaluating and generating bounding box proposals with or without the supervision of ground-truth part annotations. Based on off-the-shelf object proposals \cite{ss}, part detectors are learned in \cite{partbased} by clustering subregions of object proposals. In \cite{racnn}, a hierarchical three-level region attention mechanism is proposed that is able to attend to discriminative regions, where region discrimination is measured by classification probability. Multiple part attention maps are generated by clustering and weighting the spatially-correlated channels of the convolutional feature maps in \cite{macnn}.

There exist FGVC methods \cite{cnnoffshelf,deepoptimized,metriclearning,deepmetric,bilinear,deepimage} that process a whole image instead of local parts, yet their results are generally worse than those of part based methods. Another line of methods push the state of the art by identifying and leveraging auxiliary data beyond the ImageNet. In particular, the method of \cite{l-bird} sets an astonishing baseline on CUB-200-2011 simply by pre-training a standard deep model using a huge auxiliary set of web images that are obtained by using subordinate categories of bird as search keywords; note that such obtained auxiliary images are quite noisy in terms of their category labels.
Xie \textit{et al.} \cite{hyperclass} propose to augment the fine-grained data with a large number of auxiliary images labeled by hyper-classes; these hyper-classes are some attributes that can be annotated more easily than the fine-grained category labels, so that a large number of images labeled with attributes can be easily acquired; by joint training the model for hyper-class classification and FGVC, the performance of FGVC gets improved. Instead of searching for more semantically relevant auxiliary data from the Internet, Ge and Yu \cite{jointfinetune} propose to refine the ImageNet images by comparing them with those in the training set of the target FGVC task, using low-level features (e.g., the Gabor filter responses); such a refined ImageNet is then used to jointly train a model with training images of the FGVC task.

All the above methods use auxiliary data either to pre-train a model, or to jointly train a model with training images of the target FGVC task. In contrast, our proposed MetaFGNet uses a regularized meta-learning objective that can make full use of the auxiliary data, while at the same time making the obtained model optimal for a further adaptation to the target FGVC task. We also compare our training objective with that of joint training technically in Section \ref{SecMetaFGNet} and empirically in Section \ref{SecExp}.

\noindent\textbf{Few-shot learning via meta learning} Meta learning aims to learn experience from history and adapt to new tasks with the help of history knowledge. Few-shot learning is one of its applications. \cite{siamese} trains a siamese neural network for the task of verification, which is to identify whether input pairs belong to the same class; once the verification model is trained, it can be used for few- or one-shot learning by calculating the similarity between the test image and the labelled images. \cite{matching} realizes few-shot learning with a neural network which is augmented with external memories; it uses two embeddings to map the images to feature space and the classification is obtained by measuring the cosine distance in the feature space; the embedding of the test images can be modified by the whole support set through a LSTM attention module, which makes the model utilize the support set more reasonably and effectively. In \cite{metalstm}, SGD is replaced by a meta-LSTM that can learn an update rule for training networks. Finn \textit{et al.} \cite{maml} propose a meta learning method termed MAML, which trains a meta model in a multi-task fashion. Different from the problem setting of MAML, which learns meta models from a set of training tasks that are independent of the target task, our proposed MetaFGNet involves directly the target task into the training objective.

\section{The proposed MetaFGNet}
\label{SecMetaFGNet}

For a target FGVC of interest, suppose we have training data ${\cal{T}} = \{ (\mathbf{x}_i^t, \mathbf{y}_i^t) \}_{i=1}^{|{\cal{T}}|}$, where each pair of $\mathbf{x}_i^t$ and $\mathbf{y}_i^t$ represents an input image and its one-hot vector representation of class label. We also assume that a set of auxiliary data (e.g., the ImageNet) is available that contains images different from (but possibly semantically related to) the target data ${\cal{T}}$. Denote the auxiliary data as ${\cal{S}} = \{ (\mathbf{x}_i^s, \mathbf{y}_i^s) \}_{i=1}^{|{\cal{S}}|}$. As illustrated in Figure \ref{fig:MetaFGNet}, our proposed MetaFGNet is based on a deep neural network consisting of two parallel classifiers of fully-connected (FC) layers that share a common base network. The two classifiers are respectively used for ${\cal{T}}$ and ${\cal{S}}$. We correspondingly denote parameters of the two classifiers as $\mathbf{\theta}_c^s$ and $\mathbf{\theta}_c^t$, and denote those of the base network collectively as $\mathbf{\theta}_b$, which contains parameters of layer weights and bias. For ease of subsequent notations, we also denote the parameters of target and source model as $\mathbf{\theta}^t = (\mathbf{\theta}_b, \mathbf{\theta}_c^t)$ and $\mathbf{\theta}^s = (\mathbf{\theta}_b, \mathbf{\theta}_c^s)$ respectively.

In machine learning, ${\cal{T}}$ is usually sampled i.i.d. from an underlying (unknown) distribution ${\cal{D}}$. To learn a deep model $\mathbf{\theta}^t$, one may choose an appropriate loss function $L(\mathbf{x}^t, \mathbf{y}^t; \mathbf{\theta}^t)$, and minimize the following expected loss over ${\cal{D}}$
\begin{eqnarray}\label{EqnExpectedTargetLoss}
\min_{\mathbf{\theta}^t} \mathbf{E}_{ (\mathbf{x}^t, \mathbf{y}^t) \sim {\cal{D}} } \left[ L(\mathbf{x}^t, \mathbf{y}^t; \mathbf{\theta}^t) \right] .
\end{eqnarray}
In practice, however, minimizing the above objective is infeasible since the underlying distribution ${\cal{D}}$ is unknown. As an alternative, one chooses to minimize the following empirical loss to learn $\mathbf{\theta}^t$
\begin{eqnarray}\label{EqnEmpiricalTargetLoss}
\min_{\mathbf{\theta}^t} \frac{1}{|{\cal{T}}|} L\left({\cal{T}}; \mathbf{\theta}^t \right) = \frac{1}{|{\cal{T}}|} \sum_{i=1}^{|{\cal{T}}|} L\left( \mathbf{x}_i^t, \mathbf{y}_i^t; \mathbf{\theta}^t \right) .
\end{eqnarray}
As discussed in Section \ref{SecIntro}, fine-grained classification bears problem characteristics of few-shot learning, and its training set ${\cal{T}}$ is usually too small to well represent the underlying distribution ${\cal{D}}$. Thus directly minimizing the empirical loss (\ref{EqnEmpiricalTargetLoss}) causes severe overfitting. In the literature of fine-grained classification, this issue is usually addressed by pre-training the model $\mathbf{\theta}^t$ using an auxiliary set of data ${\cal{S}}$ (e.g., the ImageNet), and then fine-tuning it using ${\cal{T}}$. Note that this strategy alleviates overfitting in two aspects: (1) pre-training gives the model a good initialization that has learned (generic) semantic knowledge from ${\cal{S}}$; and (2) fine-tuning itself reduces overfitting via early stop of training. In other words, one may understand the strategy of fine-tuning as imposing \textit{implicit regularization} on the learning of $\mathbf{\theta}^t$. Alternatively, one may apply \textit{explicit regularization} to (\ref{EqnEmpiricalTargetLoss}), resulting in the following general form of regularized loss minimization
\begin{eqnarray}\label{EqnGeneralReguLoss}
\min_{\mathbf{\theta}^t} \frac{1}{|{\cal{T}}|} L\left({\cal{T}}; \mathbf{\theta}^t \right) + R(\mathbf{\theta}^t) .
\end{eqnarray}
The auxiliary set ${\cal{S}}$ can be used as an instantiation of the regularizer, giving rise to the following \textit{joint training method}
\begin{eqnarray}\label{EqnAuxDataReguLoss}
\min_{\mathbf{\theta}_b, \mathbf{\theta}_c^t, \mathbf{\theta}_c^s} \frac{1}{|{\cal{T}}|} L\left( {\cal{T}}; \mathbf{\theta}_b, \mathbf{\theta}_c^t \right) + \frac{1}{|{\cal{S}}|} R\left( {\cal{S}}; \mathbf{\theta}_b, \mathbf{\theta}_c^s \right) ,
\end{eqnarray}
where regularization is only imposed on parameters $\mathbf{\theta}_b$ of the base network. By leveraging ${\cal{S}}$, the joint training method (\ref{EqnAuxDataReguLoss}) could be advantageous over fine-tuning since network training has a chance to converge to a more mature, but not overfitted, solution. Based on a similar deep architecture as in Figure \ref{fig:MetaFGNet}, the joint training method (\ref{EqnAuxDataReguLoss}) is used in a recent work of fine-grained classification \cite{jointfinetune}. The choice of the auxiliary set ${\cal{S}}$ also matters in (\ref{EqnAuxDataReguLoss}). Established knowledge from the literature of transfer learning \cite{TransferLearningSurvey} suggests that ${\cal{S}}$ should ideally have similar distribution of feature statistics as that of ${\cal{T}}$, suggesting that a refinement of ${\cal{S}}$ could be useful for better regularization.

\subsection{A meta-learning objective for MetaFGNet}

Inspired by recent meta learning methods \cite{maml,matching,metalstm} that learn a meta model from a set of training few-shot learning tasks, we propose in this paper a meta-learning objective for the target fine-grained classification task ${\cal{T}}$. Instead of using the loss $L({\cal{T}}; \mathbf{\theta}^t)$ directly as in (\ref{EqnGeneralReguLoss}), the meta-learning objective is to \textit{guide the optimization of $\mathbf{\theta}^t$} so that the obtained $\mathbf{\theta}^t$ can fast adapt to the target task via a second process of fine-tuning. Suppose the fine-tuning process achieves
\begin{eqnarray}\label{EqnGeneralFineTuneUpt}
\mathbf{\theta}^t \leftarrow \mathbf{\theta}^t + \triangle(\mathbf{\theta}^t) ,
\end{eqnarray}
where $\triangle(\mathbf{\theta}^t)$ denotes the amount of parameter update. The problem nature of few-shot learning suggests that fine-tuning should be a fast process: a small number of (stochastic) gradient descent steps may be enough to learn effectively from ${\cal{T}}$, and taking too many steps may result in overfitting. One-step gradient descent can be written as
\begin{eqnarray}\label{EqnOneStepFineTuneUpt}
\triangle(\mathbf{\theta}^t) = - \eta \frac{1}{|{\cal{T}}|} \nabla_{\mathbf{\theta}^t} L({\cal{T}}; \mathbf{\theta}^t) ,
\end{eqnarray}
where $\eta$ denotes the step size. Based on (\ref{EqnOneStepFineTuneUpt}), we write our proposed \textit{regularized meta-learning objective} for fine-grained classification as
\begin{eqnarray}\label{EqnMetaObjWithOneStepUpt}
\min_{\mathbf{\theta}_b, \mathbf{\theta}_c^t, \mathbf{\theta}_c^s} \frac{1}{|{\cal{T}}|} L\left( {\cal{T}}; \mathbf{\theta}^t - \eta \frac{1}{|{\cal{T}}|} \nabla_{\mathbf{\theta}^t} L({\cal{T}}; \mathbf{\theta}^t) \right) + \frac{1}{|{\cal{S}}|} R\left( {\cal{S}}; \mathbf{\theta}^s \right) .
\end{eqnarray}
Our proposed meta-learning objective can also be explained from the perspective of reducing effective model capacity, and can thus achieve additional alleviation of overfitting apart from the effect of the regularizer $R({\cal{S}}; \mathbf{\theta}_b, \mathbf{\theta}_c^s)$, in which the regularization is achieved by base parameters updating from auxiliary data.

{\noindent\bf{Remarks}} Both our proposed MetaFGNet and the meta-learning methods \cite{maml,metalstm} contain loss terms of meta-learning, which guide the trained model to be able to fast adapt to a target task. We note that our method is for a problem setting different from those of \cite{maml,metalstm}, and consequently is the objective (\ref{EqnMetaObjWithOneStepUpt}): they learn meta models from a set of training tasks and subsequently use the learned meta model for a new target task; here training set ${\cal{T}}$ of the target task is directly involved in the main learning objective.

\subsection{Training algorithm}

Solving the proposed objective (\ref{EqnMetaObjWithOneStepUpt}) via stochastic gradient descent (SGD) involves computing gradient of a gradient for the first term, which can be derived as
\begin{eqnarray}\label{EqnHessiam_vector}
\nabla_{\mathbf{\theta}^{t'}} \frac{1}{|{\cal{T}}_j|} L\left( {\cal{T}}_j;\mathbf{\theta}^{t'}\right) \left[\mathbf{I} - \eta \frac{1}{|{\cal{T}}_i|} \left(\frac{\partial^{2} L({\cal{T}}_i; \mathbf{\theta}^t)}{\partial ({\mathbf{\theta}^{t}})^{2}} \right)\right] ,
\end{eqnarray}
where ${\cal{T}}_i$ and ${\cal{T}}_j$ denote mini-batches of ${\cal{T}}$, and $\mathbf{\theta}^{t'} = \mathbf{\theta}^{t} - \eta \frac{1}{|{\cal{T}}_i|} \nabla_{\mathbf{\theta}^t} L({\cal{T}}_i; \mathbf{\theta}^t)$. Hessian matrix is involved in (\ref{EqnHessiam_vector}), computation of which is supported by modern deep learning libraries \cite{TensorFlow,pytorch}. In this work, we adopt the Pytorch \cite{pytorch} to implement (\ref{EqnHessiam_vector}), whose empirical computation time is about 0.64s per iteration (batchsize = 32) when training MetaFGNet on a GeForce GTX 1080 Ti GPU. Training of MetaFGNet is given in Algorithm \ref{alg:metafgnet}.

\begin{algorithm}[htb]
\caption{Training algorithm for MetaFGNet}
\label{alg:metafgnet}
	\begin{algorithmic}[1]
	\Require ${\cal{T}}$: target train data; ${\cal{S}}$: auxiliary train data
	\Require $\eta, \alpha$: hyperparameters of step size
	\State initialize $\mathbf{\theta}_b$, $\mathbf{\theta}_c^t$, $\mathbf{\theta}_c^s$
	\While {not done}
		\State Sample mini-batches ${\cal{T}}_i$, ${\cal{S}}_i$ from ${\cal{T}}$, ${\cal{S}}$
		\State Evaluate: \mbox{} \par $[\triangle(\theta_b;{{\cal{S}}_i}),\triangle(\theta_c^s;{{\cal{S}}_i})] = \frac{1}{|{\cal{S}}_i|} \nabla_{\mathbf{\theta}^s} R\left( {\cal{S}}_i; \mathbf{\theta}^s \right)$ \mbox{} \par $[\triangle(\theta_b;{{\cal{T}}_i}),\triangle(\theta_c^t; {{\cal{T}}_i})] = \frac{1}{|{\cal{T}}_i|} \nabla_{\mathbf{\theta}^t} L({\cal{T}}_i; \mathbf{\theta}^t)$
		\State  Compute adapted parameters with SGD: \mbox{} \par $\mathbf{\theta}^{t'} = \mathbf{\theta}^{t} - \eta \frac{1}{|{\cal{T}}_i|} \nabla_{\mathbf{\theta}^t} L({\cal{T}}_i; \mathbf{\theta}^t)$
		\State Sample another mini-batch ${\cal{T}}_j$ from ${\cal{T}}$
		\State Evaluate: \mbox{} \par $[\triangle(\theta_b;{{\cal{T}}_j}),\triangle(\theta_c^t;{{\cal{T}}_j})] = \nabla_{\mathbf{\theta}^{t'}} \frac{1}{|{\cal{T}}_j|} L\left( {\cal{T}}_j;\mathbf{\theta}^{t'}\right) \left[\mathbf{I} - \eta \frac{1}{|{\cal{T}}_i|} \left(\frac{\partial^{2} L({\cal{T}}_i; \mathbf{\theta}^t)}{\partial ({\mathbf{\theta}^{t}})^{2}} \right)\right]$
		\State Update: \mbox{} \par $\theta_b \gets \theta_b - \alpha [\triangle(\theta_b;{{\cal{S}}_i}) + \triangle(\theta_b;{{\cal{T}}_j})]$  \mbox{} \par $\theta_c^t \gets \theta_c^t - \alpha \triangle(\theta_c^t;{{\cal{T}}_j})$ \mbox{} \par $\theta_c^s \gets \theta_c^s - \alpha \triangle(\theta_c^s;{{\cal{S}}_i}) $
	\EndWhile
	\end{algorithmic}
\end{algorithm}

\section{Sample selection of auxiliary data using the proposed MetaFGNet}
\label{SecSampleSelection}

Established knowledge from domain adaptation suggests that the auxiliary set ${\cal{S}}$ should ideally have a similar distribution of feature statistics as that of the target set ${\cal{T}}$. This can be achieved either via transfer learning \cite{TransferLearningSurvey}, or via selecting/refining samples of ${\cal{S}}$. In this work, we take the second approach and propose a simple sample selection scheme that is naturally supported by our proposed MetaFGNet (and in fact by any deep models with a two-head architecture as in Figure \ref{fig:MetaFGNet}).

Given a trained MetaFGNet, for each auxiliary sample $\mathbf{x}^s$ from ${\cal{S}}$, we compute through the network to get two \textit{prediction vectors} $\mathbf{z}_s^s$ and $\mathbf{z}_t^s$, which are respectively the output vectors of the two classifiers (before the softmax operation) for the source and target tasks. Length of $\mathbf{z}_s^s$ (or $\mathbf{z}_t^s$) is essentially equal to category number of the source task (or that of the target task). To achieve sample selection from the auxiliary set ${\cal{S}}$, we take the approach of assigning a score to each $\mathbf{x}^s$ and then ranking scores of all auxiliary samples. The score of $\mathbf{x}^s$ is computed as follows: we first set negative values in $\mathbf{z}_s^s$ and $\mathbf{z}_t^s$ as zero; we then concatenate the resulting vectors and apply L2 normalization, producing $\tilde{\mathbf{z}}^s = [\tilde{\mathbf{z}}_s^{s\top}, \tilde{\mathbf{z}}_t^{s\top}]^{\top}$; we finally compute the score for $\mathbf{x}^s$ as
\begin{eqnarray}\label{Eqntwoscores}
O^s = \tilde{\mathbf{z}}_t^{s\top} \cdot \mathbf{1} ,
\end{eqnarray}
where $\mathbf{1}$ represents a vector with all entry values of $1$. A specified ratio of top samples can be selected from ${\cal{S}}$ and form a new set of auxiliary data. Rationale of the above scheme lies in that auxiliary samples that are more semantically related to the target task would have higher responses in the target classifier, and consequently would have higher values in $\tilde{\mathbf{z}}_t^{s\top}$.

Experiments in Section \ref{SecExp} show that such a sample selection scheme is effective to select images that are semantically more related to the target task and improve performance of fine-grained classification. Some high-scored and low-scored samples in the auxiliary data are also visualized in Figure \ref{Fig.abandonedImage}.


\section{Experiments}
\label{SecExp}

\subsection{Datasets and implementation details}

\noindent\textbf{CUB-200-2011} The CUB-200-2011 dataset \cite{cub2002011} contains 11,788 bird images. There are altogether 200 bird species and the number of images per class is about 60. The significant variations in pose, viewpoint, and illumination inside each class make this task very challenging. We adopt the publicly available split \cite{cub2002011}, which uses nearly half of the dataset for training and the other half for testing.
\noindent\textbf{Stanford Dogs} The Stanford Dogs dataset \cite{stanforddog} contains 120 categories of dogs. There are 12,000 images for training and 8,580 images for testing. This dataset is also challenging due to small inter-class variation, large intra-class variation, and cluttered background.

\noindent\textbf{ImageNet Subset} The ImageNet Subset contains all categories of the original ImageNet \cite{imagenet} except the 59 categories of bird species, providing more realistic auxiliary data for CUB-200-2011 \cite{cub2002011}. Note that almost all existing methods on CUB-200-2011 use the whole ImageNet dataset as the auxiliary set.

\noindent\textbf{L-Bird Subset} The original L-Bird dataset \cite{l-bird} contains nearly 4.8 million images which are obtained by searching images of a total of 10,982 bird species from the Internet. The dataset provides urls of these images, and by the time of our downloading the dataset, we only manage to get 3.3 million images from effective urls. To build the dataset of L-Bird Subset, we first choose bird species/classes out of the total 10,982 ones whose numbers of samples are beyond 100; we then remove all the 200 bird classes that are already used in CUB-200-2011, since the L-Bird Subset will be used as an auxiliary set for CUB-200-2011; we finally hold out 1$\%$ of the resulting bird images as a validation set, following the work of \cite{l-bird}. The final auxiliary L-Bird Subset contains 3.2 million images.

\noindent\textbf{Remarks on the used datasets} We use the ImageNet Subset, the ImageNet ILSVRC 2012 training set \cite{imagenet}, or the L-Bird Subset as the set of auxiliary data for CUB-200-2011, and use the ImageNet ILSVRC 2012 training set as the set of auxiliary data for Stanford Dogs.

\noindent\textbf{Implementation details} Many existing convolutional neural networks (CNNs), such as AlexNet \cite{alexnet}, VGGNet \cite{vgg}, or ResNet \cite{resnet}, can be used as backbone of our MetaFGNet. In this work, we use the pre-activation version of the 34-layer ResNet \cite{resnet} in our experiments, \emph{which can achieve almost identical performance on ImageNet with a batch normalization powered VGG16}. To adapt any of them for MetaFGNet, we remove its last fully-connected (FC) layer and keep the remaining ones as the base network of MetaFGNet, which are shared by the auxiliary and target data as illustrated in Figure \ref{fig:MetaFGNet}. Two parallel FC layers of classifiers are added on top of the base network which are respectively used for the meta-learning objective of the target task and the regularization loss of the auxiliary task. The MetaFGNet adapted from the 34-layer ResNet is used for both the ablation studies and the comparison with the state of the art. For a fair comparison with existing FGVC methods, base network is pre-trained on ImageNet for all experiments reported in this paper. When using ImageNet Subset or ImageNet as the auxiliary data, we start from the 60th epoch pre-trained model, mainly for a quick comparison with baseline methods. When using L-Bird Subset as the auxiliary data, we employ the released pre-trained model from \cite{resnet}. The architectural design of our MetaFGNet is straightforward and simple; in contrast, most of existing FGVC methods \cite{spda,racnn,hsnet} adopt more complicated network architectures in order to exploit discrimination of local parts with or without use of ground-truth part annotations.

During each iteration of SGD training, we sample one mini-batch of auxiliary data for the regularization loss, and two mini-batches of target data for the meta-learning loss (cf. Algorithm \ref{alg:metafgnet} for respective use of the two mini-batches). Each mini-batch includes 256 images. We do data augmentation on these images according to \cite{resnet}.
In experiments using ImageNet Subset or ImageNet as the auxiliary data, the learning rate ($\alpha$ in Algorithm \ref{alg:metafgnet}) starts from 0.1 and is divided by 10 after every 10 epochs; we set momentum as 0.9 and weight decay as 0.0001; the meta learning rate ($\eta$ in Algorithm \ref{alg:metafgnet}) starts from 0.01 and is divided by 10 after every 10 epochs, in order to synchronize with the learning rate; the experiments end after 30 training epochs, which gives a total of the same 90 epochs as that of a pre-trained model. When using L-Bird Subset as the auxiliary data, the experiments firstly fine-tune an ImageNet pre-trained model on L-Bird Subset for 32 epochs, and then train our MetaFGNet for 8 epochs starting from the 24th epoch fine-tuned model; the learning rate and meta learning rate are divided by 10 respectively after 4 and 6 epochs; other settings are the same as in experiments using ImageNet or ImageNet Subset as the auxiliary data. Given parameters of such trained MetaFGNets and re-initialized target classifiers, we fine-tune $(\mathbf{\theta}_b, \mathbf{\theta}_c^t)$ of them on the target data for another 160 epochs, which is the same for all comparative methods. We do sample selection from the auxiliary data as the way described in Section \ref{SecSampleSelection}, \textit{using the trained MetaFGNets before fine-tuning}. For the auxiliary sets of ImageNet Subset, ImageNet and L-Bird Subset, we respectively use $50\%$, $6\%$, and $80\%$ of their samples as the selected top samples. Note that such ratios are empirically set and are suboptimal. After sample selection, we use the remained auxiliary samples to form a new auxiliary set, and train and fine-tune a MetaFGNet again from the MetaFGNet that have been trained using the original auxiliary datasets.

\subsection{Comparison with alternative baselines}
\label{Sec_baselines}

The first baseline method (referred as ``Fine-tuning'' in tables reported in this subsection) simply fine-tunes on the target dataset a model that has been pre-trained on the ImageNet Subset or ImageNet, of which the latter is typically used in existing FGVC methods. The second baseline (referred as ``Joint training'' in tables reported in this subsection) uses a joint training approach of the objective (\ref{EqnAuxDataReguLoss}).  The third baseline (referred as ``Fine-tuning L-Bird Subset'' in tables reported in this subsection) firstly fine-tunes the ImageNet pre-trained 34-layer ResNet model on L-bird Subset, and then fine-tunes the resulting model on CUB-200-2011. Experiments in this subsection are based on a MetaFGNet adapted from the 34-layer ResNet, for which we refer to it as ``MetaFGNet''. The baselines of Fine-tuning and Fine-tuning L-Bird Subset use half of the MetaFGNet that contains parameters of $(\mathbf{\theta}_b, \mathbf{\theta}_c^t)$. The baseline of Joint training uses the same MetaFGNet structure as our method does.

Tables \ref{table:ImageNet_subset} and \ref{table:ImageNet_Lbird} report these controlled experiments on the CUB-200-2011 dataset \cite{cub2002011}. Using ImageNet Subset or ImageNet as the set of auxiliary data, Fine-tuning gives baseline classification accuracies of $73.5\%$ and $76.8\%$ respectively; the result of Joint training is better than that of Fine-tuning, suggesting the usefulness of the objective (\ref{EqnAuxDataReguLoss}) for FGVC tasks - note that a recent method \cite{jointfinetune} is essentially based on this objective. Our proposed MetaFGNet with regularized meta-learning objective (\ref{EqnMetaObjWithOneStepUpt}) achieves a result better than that of Joint training. Our proposed sample selection scheme further improves the results to $75.3\%$ and $80.3\%$ respectively, thus justifying the efficacy of our proposed method. When using L-Bird Subset as the auxiliary set, our method without sample selection improves the result to $87.2\%$, showing that a better auxiliary set is essential to achieve good performance on FGVC tasks. Note that L-Bird Subset does not contain the 200 bird species of the CUB-200-2011 dataset. Our method with sample selection further improves the result to $87.6\%$, confirming the effectiveness of our proposed scheme. Samples of the selected images and abandoned images from three auxiliary datasets are also shown in Section \ref{SeeImage}.

\begin{table}[h]
\caption{Comparative studies of different methods on the CUB-200-2011 dataset \cite{cub2002011}, using  ImageNet Subset as the auxiliary set.  Experiments are based on networks adapted from a 34-layer ResNet.}
\label{table:ImageNet_subset}
\begin{center}
\begin{tabular}{p{6.5cm}p{3cm}p{2.2cm}<{\centering}} 
\hline
\hline
Methods                               & Auxiliary set   & Accuracy (\%) \\
\hline

\hline
Fine-tuning                          & ImageNet Subset & 73.5\\
\hline
Joint training w/o sample selection  & ImageNet Subset & 74.5\\
\hline
Joint training with sample selection & ImageNet Subset & 75.0\\
\hline
MetaFGNet w/o sample selection       & ImageNet Subset & 75.0\\
\hline
MetaFGNet with sample selection      & ImageNet Subset & 75.3\\

\hline
\hline
\end{tabular} 
\end{center}
\end{table}
\vspace{-0.5cm}
\begin{table}[h]
\caption{Comparative studies of different methods on the CUB-200-2011 dataset \cite{cub2002011}, using  ImageNet \cite{imagenet} or L-Bird Subset as the auxiliary set.}
\label{table:ImageNet_Lbird}
\begin{center}
\begin{tabular}{p{6.5cm}p{3cm}p{2.2cm}<{\centering}} 
\hline
\hline
Methods                               & Auxiliary set   & Accuracy (\%) \\
\hline

\hline
Fine-tuning                          & ImageNet        & 76.8\\
\hline
Joint training w/o sample selection  & ImageNet        & 78.8\\
\hline
Joint training with sample selection & ImageNet        & 79.4\\
\hline
MetaFGNet w/o sample selection       & ImageNet        & 79.5 \\
\hline
MetaFGNet with sample selection      & ImageNet        & 80.3 \\

\hline

\hline

Fine-tuning L-Bird Subset               & L-Bird Subset   & 86.2 \\
\hline
MetaFGNet w/o sample selection       & L-Bird Subset   & 87.2\\
\hline
MetaFGNet with sample selection      & L-Bird Subset   & 87.6\\
\hline
\hline
\end{tabular} 
\end{center}
\end{table}

\begin{figure}[t]
\begin{center}
\includegraphics[width=0.48\linewidth]{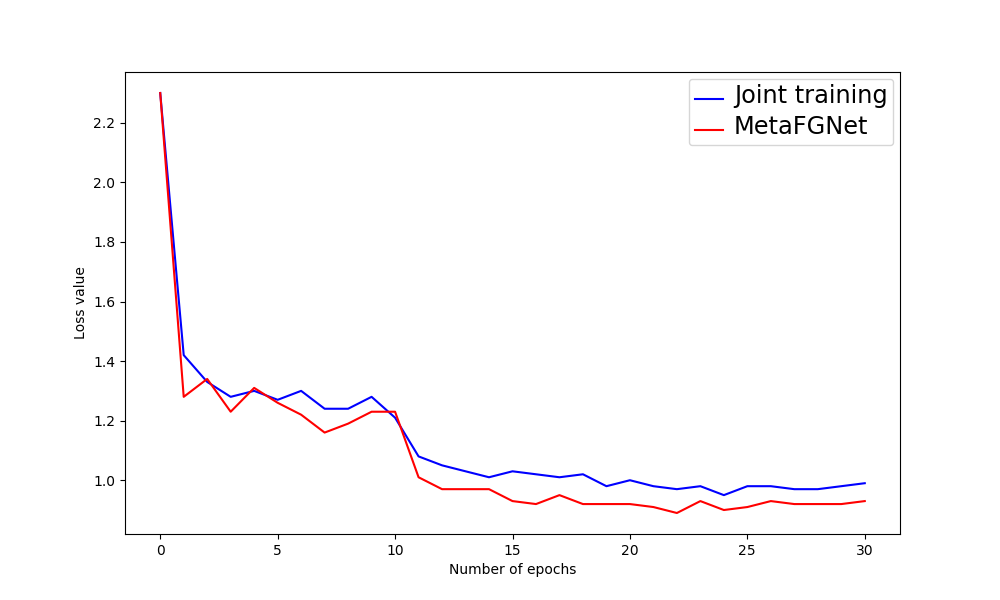} \hfill  \includegraphics[width=0.48\linewidth]{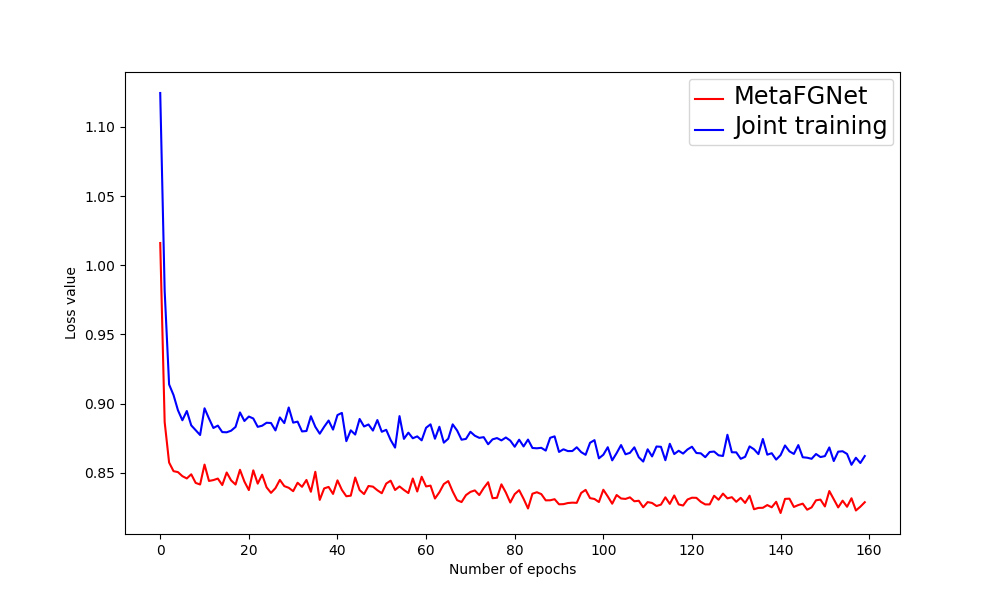} \\ 
\end{center}
\caption{ Left: training loss curves of MetaFGNet and Joint training. Right: fine-tuning loss curves of 
MetaFGNet and Joint training. The auxiliary and target datasets are ImageNet and CUB-200-2011 respectively. MetaFGNet and Joint training models are adapted from the 34-layer ResNet. }
\label{fig:training loss}
\end{figure}

In Figure \ref{fig:training loss}, we also plot the training loss curves, using both the ImageNet auxiliary data and the target CUB-200-2011 data, of our method and Joint training, and also their fine-tuning loss curves on the target data. Figure \ref{fig:training loss} shows that our MetaFGNet converges to a better solution that supports a better fine-tuning than Joint training does. 

\subsection{Results on the CUB-200-2011}
\label{Sec_CUB_Results}


We use the MetaFGNet adapted from a 34-layer ResNet
to compare with existing methods on CUB-200-2011 \cite{cub2002011}. The most interesting comparison is with the methods \cite{cnnoffshelf,deepoptimized,metriclearning,deepmetric,bilinear,deepimage,l-bird} that focus on learning from the whole bird images. In contrast, part based methods \cite{partbased,deepfilter,racnn,macnn,partrcnn,spda,domainadaptation,hsnet} enjoy a clear advantage by exploiting discrimination of local parts either in a weakly supervised manner, or in a supervised manner using ground-truth part annotations. 
Table \ref{table:bird_result} also shows that our method with L-Bird Subset as auxiliary data outperforms all existing methods even  when they use ground-truth part annotations. We also construct our MetaFGNet based on the popular VGGNet. Using L-Bird Subset as the auxiliary set, our MetaFGNet with sample selection gives an accuracy of 87.5\%, which also confirms the efficacy of our proposed method.  

\begin{table}[h]
\caption{ Comparison of different methods on the CUB-200-2011 dataset \cite{cub2002011}. Methods of `part supervised' use ground-truth part annotations of bird images. Methods of `part aware' detect discriminative local parts in a weakly supervised manner. Result of \cite{l-bird} is based on own implementation using the currently available L-Bird Subset. \textit{The method \cite{bilinear} generates high-dimensional, second-order features via bilinear pooling, and the result of \cite{deepimage} is from a multi-scale ensemble model; both of the methods make contributions to image based FGVC complementary to our proposed MetaFGNet.}
}
\label{table:bird_result}
\begin{center}
\begin{tabular}{p{5.1cm}p{2.8cm}p{2.5cm}p{1.5cm}<{\centering}} 
\hline
\hline
Methods                                & Auxiliary Set    & Part Supervision   & Acc. (\%)  \\

\hline

\hline

CNNaug-SVM \cite{cnnoffshelf}                          & ImageNet          & n/a                         & 61.8 \\


Deep Optimized \cite{deepoptimized}                    & ImageNet          & n/a                         & 67.1 \\

MsML \cite{metriclearning}                             & ImageNet          & n/a                         & 67.9 \\

Deep Metric \cite{deepmetric}                          & ImageNet + Web        & n/a                         & 80.7 \\

Bilinear \cite{bilinear}                               & ImageNet          & n/a                         & 84.1 \\

Deep Image \cite{deepimage}                            & ImageNet          & n/a                         & 84.9 \\

Rich Data \cite{l-bird}               & L-Bird Subset    & n/a               & {\textbf{86.2}} \\

\hline

\hline

MetaFGNet with sample selection       & ImageNet         & n/a               & 80.3 \\


MetaFGNet with sample selection    & L-Bird Subset    & n/a               & {\textbf{87.6}} \\

\hline
\hline

& & & \\

\hline
\hline
Weakly sup. \cite{partbased}          & ImageNet         & part-aware        & 80.4 \\

PDFS \cite{deepfilter}                & ImageNet         & part-aware        & 84.5\\

RA-CNN \cite{racnn}                   & ImageNet         & part-aware        & 85.3 \\


MA-CNN \cite{macnn}                  & ImageNet         & part-aware        & 86.5 \\

\hline

\hline
Part R-CNN\cite{partrcnn}             & ImageNet         & part supervised             & 73.9\\

SPDA-CNN\cite{spda}                   & ImageNet         & part supervised             & 81.0\\

Webly-sup.\cite{domainadaptation}     & ImageNet + Web   & part supervised             & 84.6\\

Hsnet\cite{hsnet}                     & ImageNet         & part supervised             & 87.5 \\

\hline
\hline
\end{tabular} 

\end{center}
\end{table}

\subsection{Results on the Stanford Dogs}


We apply the MetaFGNet to the Stanford Dogs dataset \cite{stanforddog}, using ImageNet as the auxiliary data. The used MetaFGNet is adapted from a 152-layer ResNet \cite{resnet}, which is the same as the one used in the state-of-the-art method \cite{jointfinetune}. Table \ref{table:dog_result} shows the comparative results. Our method outperforms all exiting methods with a large margin. \emph{We note that previous methods use ImageNet as the auxiliary data for the Stanford Dogs task, however, it is inappropriate because the dataset of Stanford Dogs is a subset of ImageNet.} Thus, we remove all the 120 categories of dog images from ImageNet to introduce an appropriate auxiliary dataset for the Stanford Dogs dataset \cite{stanforddog}. Based on a 34-layer ResNet \cite{resnet}, simple fine-tuning after pre-training on the resulting ImageNet images gives an accuracy of 69.3\%; our MetaFGNet with sample selection improves it to 73.2\%.
\vspace{-0.5cm}
\begin{table}[!h]
\caption{Comparison of different methods on the Stanford Dogs dataset \cite{stanforddog}.}
\label{table:dog_result}
\begin{center}
\begin{tabular}{p{5.0cm}p{2.8cm}p{2.5cm}p{1.5cm}<{\centering}}
\hline
\hline

Methods                       & Auxiliary Set      & Part Supervision & Acc. (\%) \\
\hline

\hline
Weakly sup. \cite{partbased} & ImageNet           & part-aware  &  80.4 \\
DVAN \cite{dvan}             & ImageNet          & part-aware  & 81.5 \\
\hline

\hline
MsML\cite{metriclearning}   &  ImageNet          & n/a             &  70.3  \\
MagNet\cite{metricadaptive} & ImageNet           & n/a             &  75.1 \\

Selective joint training\cite{jointfinetune} & ImageNet & n/a      &  90.3 \\
\hline

\hline
MetaFGNet with sample selection              & ImageNet           & n/a         & \textbf{96.7} \\
\hline
\hline
\end{tabular} 
\end{center}
\end{table}

\subsection{Analysis of selected and abandoned auxiliary images}
\label{SeeImage}

\begin{figure}[!htb]
\centering
\includegraphics[width=0.48\linewidth]{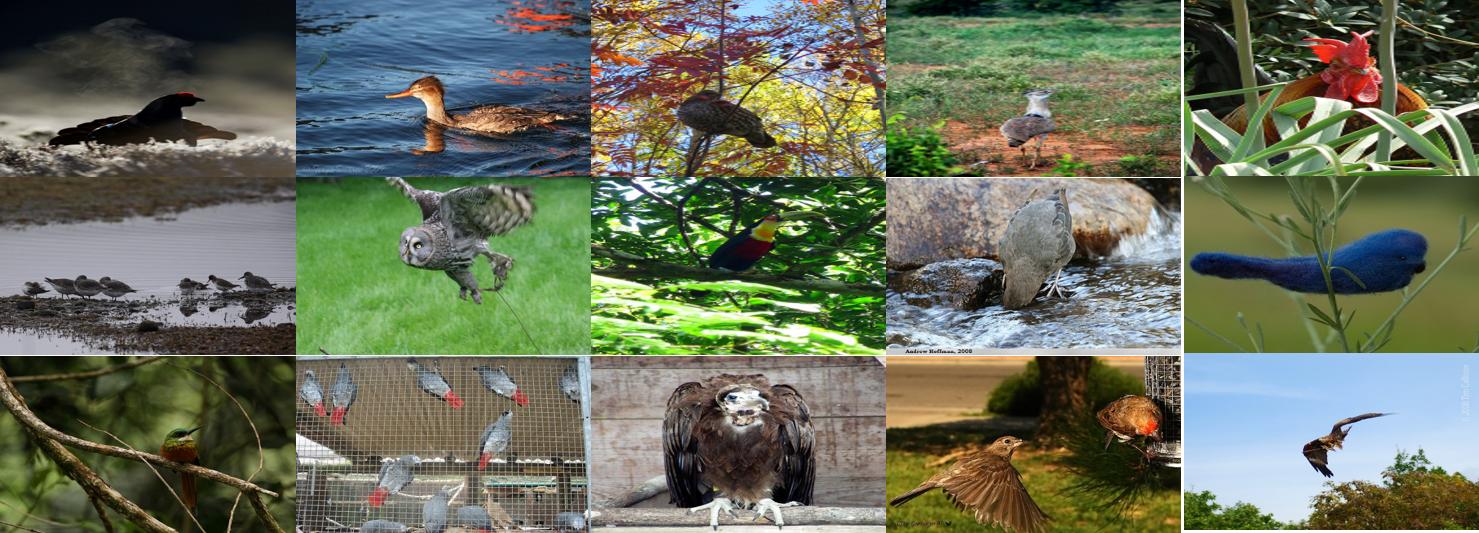} \hfill  \includegraphics[width=0.48\linewidth]{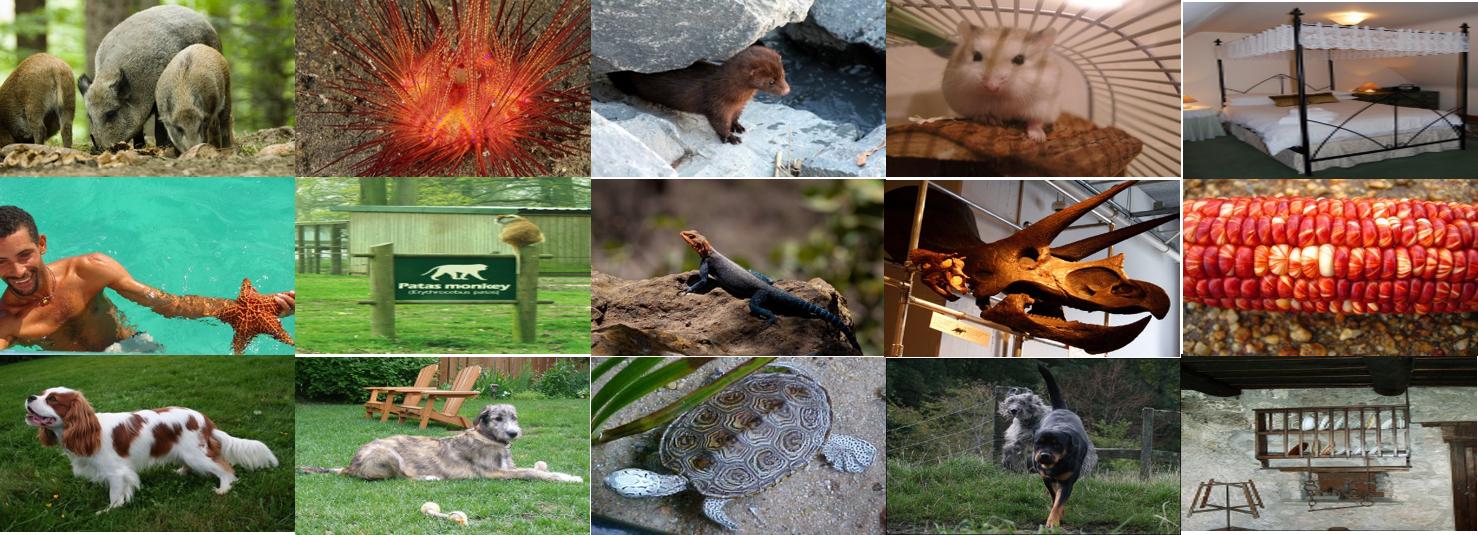} \\

(a) Top-ranked images (left) and bottom-ranked images (right) from ImageNet \\

\includegraphics[width=0.48\linewidth]{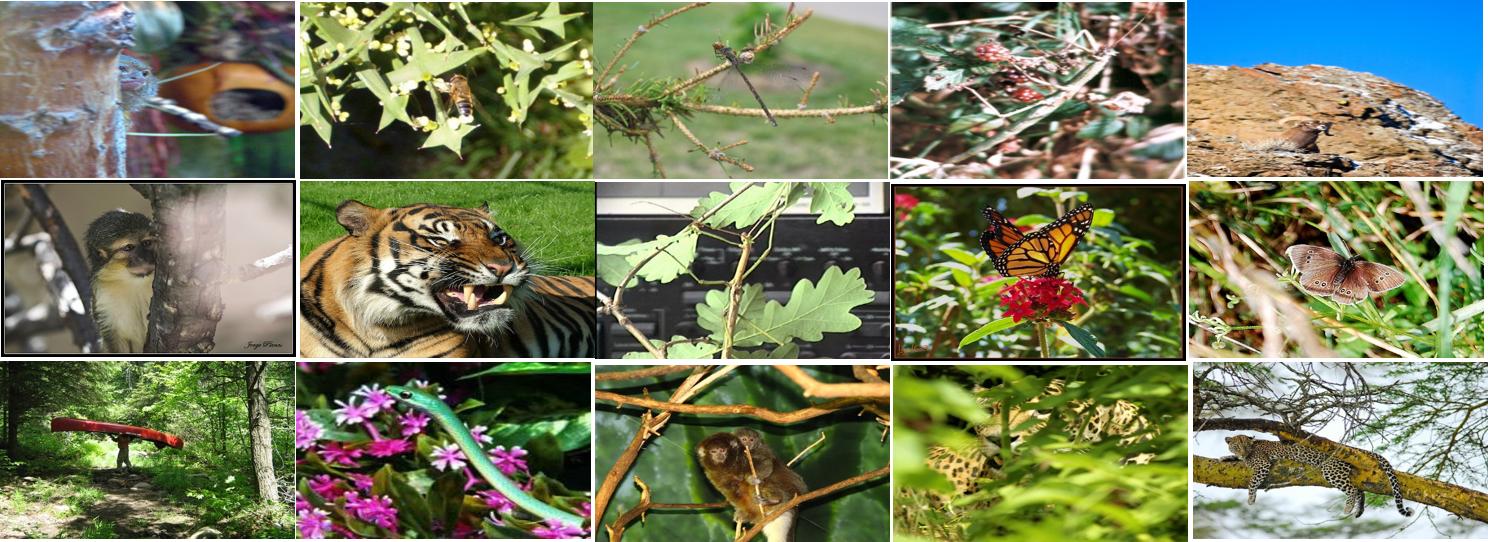} \hfill  \includegraphics[width=0.48\linewidth]{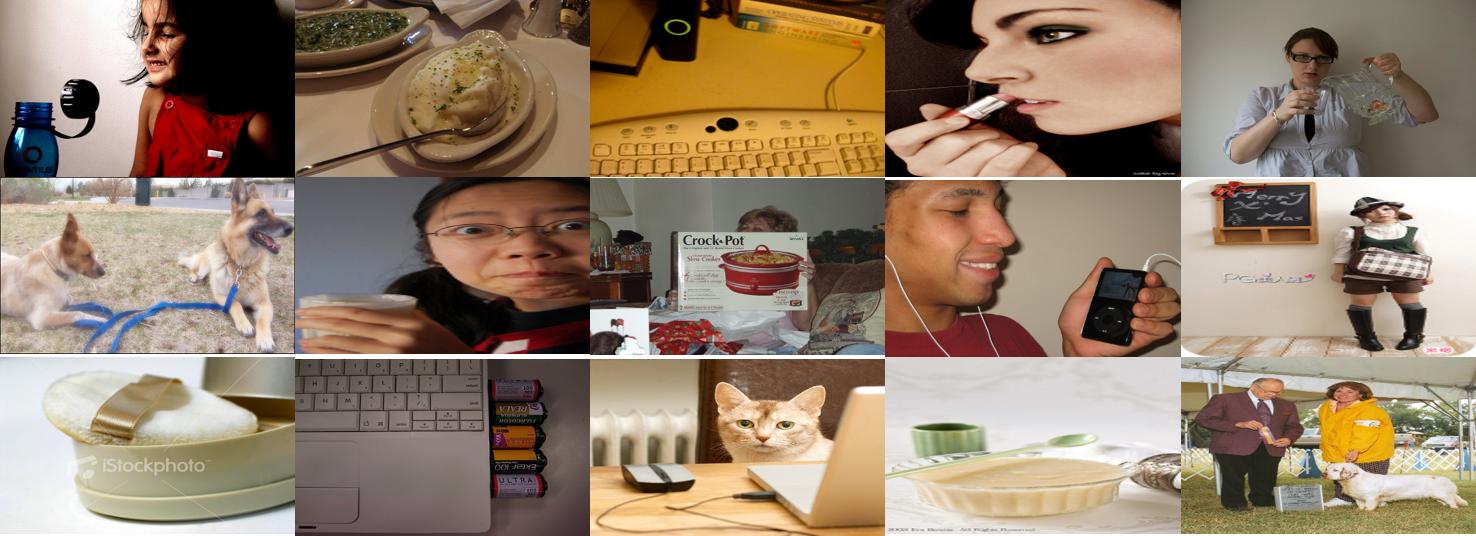} \\

(b) Top-ranked images (left) and bottom-ranked images (right) from ImageNet Subset \\

\includegraphics[width=0.48\linewidth]{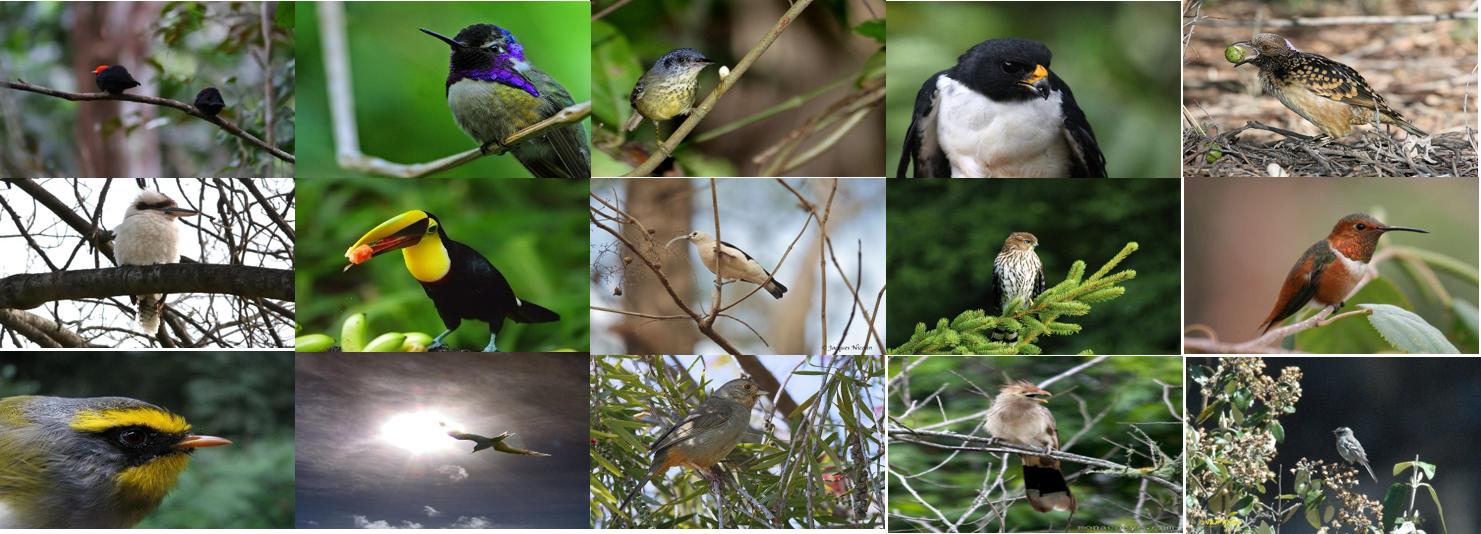} \hfill  \includegraphics[width=0.48\linewidth]{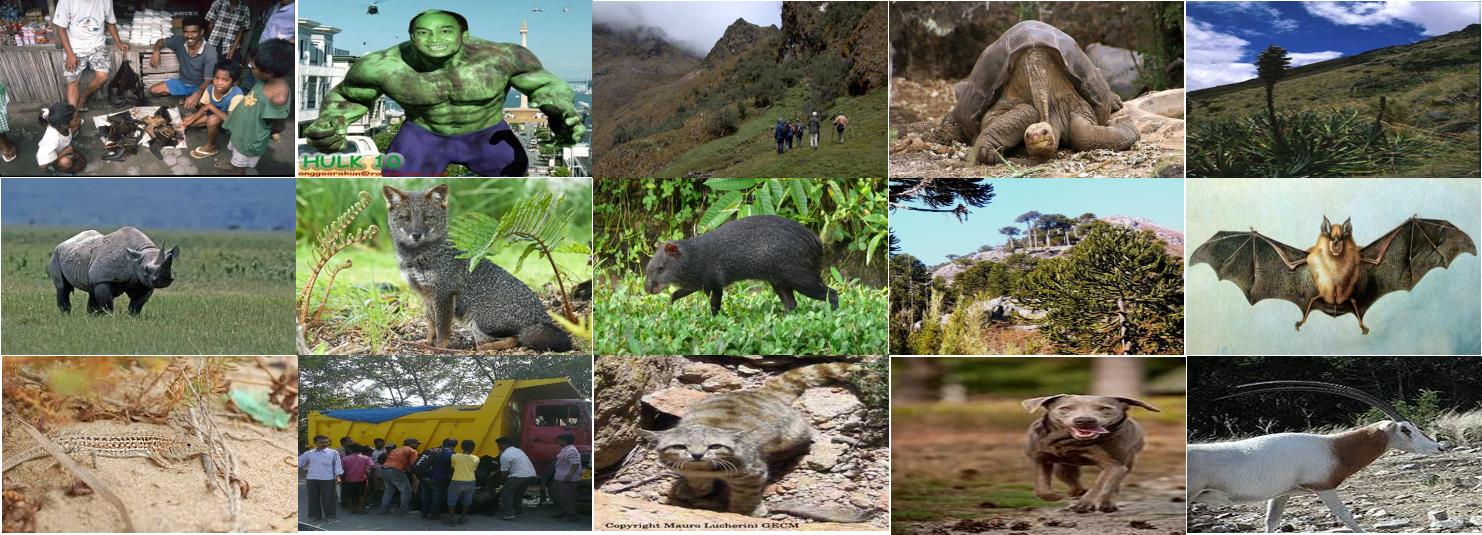} \\

(c) Top-ranked images (left) and bottom-ranked images (right) from L-Bird Subset \\

\caption{
(a) Top-ranked images and bottom-ranked images from ImageNet. (b) Top-ranked images and bottom-ranked images from ImageNet Subset. (c) Top-ranked images and bottom-ranked images from L-Bird Subset. Results are obtained by using MetaFGNet and our sample selection scheme on the CUB-200-2011 dataset \cite{cub2002011}. 
 }  \label{Fig.abandonedImage}
\end{figure}

In Fig \ref{Fig.abandonedImage}, we qualitatively visualize the selected top-ranked images from ImageNet \cite{imagenet}, ImageNet Subset, and L-Bird Subset, and also the abandoned bottom-ranked images respectively from the three auxiliary sets, when using CUB-200-2011 \cite{cub2002011} as the target data. It can be observed that for ImageNet and ImageNet Subset, images that are semantically related to the target CUB-200-2011 task are ranked top and selected by our proposed scheme; for L-Bird Subset, noisy images that are irrelevant to the target task are ranked bottom and removed. Quantitatively, when using ImageNet as the auxiliary dataset, 65.3\% of the selected auxiliary images belong to the basic-level category of ``bird''. 

%
%
%
%

\section{Conclusion}

In this paper, we propose a new deep learning model termed MetaFGNet, which is based on a novel regularized meta-learning objective that aims to guide the learning of network parameters so that they are optimal for adapting to a target FGVC task. Based on MetaFGNet, we also propose a simple yet effective scheme for sample selection from auxiliary data. Experiments on the benchmark CUB-200-2011 and Stanford Dogs datasets show the efficacy of our proposed method.

\vspace{0.2cm}
\noindent \textbf{Acknowledgment.} This work is supported in part by the Program for Guangdong Introducing Innovative and Enterpreneurial Teams (Grant No.: 2017ZT07X183), and the National Natural Science Foundation of China (Grant No.: 61771201).

%
%
%
\bibliographystyle{splncs04}
\bibliography{egbib}
\end{document}